\documentclass[11pt,a4paper,reqno]{amsart}

\usepackage[T1]{fontenc}
\usepackage{dsfont}
\usepackage{graphicx}
\usepackage{amssymb}
\usepackage{exscale}
\usepackage[breaklinks,colorlinks,citecolor=blue,linkcolor=blue]{hyperref}
\usepackage{xcolor}
\usepackage{overpic}
\usepackage[autolanguage]{numprint}
\usepackage{booktabs}
\usepackage{standalone}
\usepackage{graphicx}
\usepackage{grffile}
\usepackage{tikz}
\usetikzlibrary{arrows,backgrounds,patterns}
\usetikzlibrary{shapes,snakes}
\usepackage{mathtools}
\usepackage{enumitem}
\usepackage{import}
\usepackage{pgfplots}
\usepackage{upgreek}
\pgfplotsset{compat=newest}
\usepackage{tabularx}
\usepackage{subcaption}
\usepackage{lmodern}
\usepackage{color, soul}
\usepackage{sistyle}
\SIthousandsep{,}

\allowdisplaybreaks 

\setitemize{itemsep=5pt}

\usepackage{tikz}
\usetikzlibrary{calc}
\usetikzlibrary{shapes.geometric, arrows}

\tikzstyle{startstop} = [rectangle, rounded corners, minimum width=3cm, minimum height=1cm,text centered, draw=black, fill=green!15]
\tikzstyle{io} = [trapezium, trapezium left angle=70, trapezium right angle=110, minimum width=3cm, minimum height=1cm, text centered, draw=black, fill=blue!30, text width=5cm]
\tikzstyle{process} = [rectangle, minimum width=3cm, minimum height=1cm, text centered, draw=black, fill=gray!15, text width=5cm]
\tikzstyle{arrow} = [thick,->,>=stealth]

\newcolumntype{L}{>{\raggedright\arraybackslash\hspace{0pt}}X}
\renewcommand{\arraystretch}{1.2}

\calclayout

\newtheorem{assumption}{Assumption}

\theoremstyle{definition}
\newtheorem{remark}{Remark}

\newcommand{\myVar}{\ensuremath{\Sigma}}

\newcommand{\myGewichte}{\Omega}
\newcommand{\enbrace}[1]{\ensuremath{\left(#1\right)}} 
\newcommand{\barenbrace}[1]{\ensuremath{\left[#1\right]}} 
\newcommand{\setenbrace}[1]{\ensuremath{\left\lbrace#1\right\rbrace}} 
\newcommand{\dpart}[1]{\ensuremath{\partial_{#1}}}
\newcommand{\ddpart}[2]{\ensuremath{\partial^2_{#1 #2}}}

\newcommand{\overbar}[1]{\mkern 1.5mu\overline{\mkern-1.5mu#1\mkern-1.5mu}\mkern 1.5mu} 
\newcommand{\R}{\mathds{R}} 
\newcommand{\E}{\mathds{E}} 
\newcommand{\ND}{\mathcal{N}} 
\renewcommand{\ast}{\star}

\newcommand{\figref}[1]{\figurename~\ref{#1}}
\newcommand{\bm}[1]{#1}
\newcommand{\tdiff}[1]{\frac{\mathrm{d}#1}{\mathrm{d}t}}
\newcommand{\ddiff}[2]{\frac{\partial#1}{\partial#2}}

\usepackage[disable]{todonotes}

\begin{document}

\title[Identification of Model Uncertainty]{Identification of Model Uncertainty via Optimal Design of Experiments applied to a mechanical press}

\author[Gally]{Tristan Gally}
\address{Tristan Gally \\ Technische Universit\"at Darmstadt, Department of Mathematics, Research Group Optimization\\ Dolivostra\ss e 15, 64293 Darmstadt, Germany}
\email{gally@mathematik.tu-darmstadt.de}

\author[Groche]{Peter Groche}
\address{Peter Groche \\ Technische Universit\"at Darmstadt, Institute for Production Engineering and Forming Machines\\ Otto-Berndt-Stra\ss e 2, 
64287 
Darmstadt, Germany}
\email{groche@ptu.tu-darmstadt.de}

\author[Hoppe]{Florian Hoppe}
\address{Florian Hoppe \\ Technische Universit\"at Darmstadt, Institute for Production Engineering and Forming Machines\\ Otto-Berndt-Stra\ss e 2, 
64287 
Darmstadt, Germany}
\email{hoppe@ptu.tu-darmstadt.de}

\author[Kuttich]{Anja Kuttich}
\address{Anja Kuttich \\ Technische Universit\"at Darmstadt, Department of Mathematics, Research Group Optimization\\ Dolivostra\ss e 15, 64293 Darmstadt, Germany}
\email{kuttich@mathematik.tu-darmstadt.de\pagebreak}

\author[Matei]{Alexander Matei}
\address{Alexander Matei \\ Technische Universit\"at Darmstadt, Department of Mathematics, Research Group Optimization\\ Dolivostra\ss e 15, 64293 Darmstadt, Germany}
\email[Corresponding author]{matei@mathematik.tu-darmstadt.de}

\author[Pfetsch]{Marc E. Pfetsch}
\address{Marc E. Pfetsch \\ Technische Universit\"at Darmstadt, Department of Mathematics, Research Group Optimization\\ Dolivostra\ss e 15, 64293 Darmstadt, Germany}
\email{pfetsch@mathematik.tu-darmstadt.de}

\author[Rakowitsch]{Martin Rakowitsch}
\address{Martin Rakowitsch \\ Technische Universit\"at Darmstadt, Institute for Production Engineering and Forming Machines\\ Otto-Berndt-Stra\ss e 
2, 
64287 Darmstadt, Germany}
\email{martin.rakowitsch@gmx.de}

\author[Ulbrich]{Stefan Ulbrich}
\address{Stefan Ulbrich \\ Technische Universit\"at Darmstadt, Department of Mathematics, Research Group Optimization\\ Dolivostra\ss e 15, 64293 Darmstadt, Germany}
\email{ulbrich@mathematik.tu-darmstadt.de}


\begin{abstract}
In engineering applications almost all processes are described with the help of models. 
Especially forming machines heavily rely on mathematical models for 
control and condition monitoring. 
Inaccuracies during the modeling, manufacturing and assembly of these machines induce model uncertainty which impairs the 
controller's performance. 
In this paper we propose an approach to identify model uncertainty using parameter identification, optimal design of experiments and hypothesis testing. 
The experimental setup is characterized by optimal sensor positions such that specific model parameters can be 
determined with minimal variance. 
This allows for the computation of confidence regions in which the real parameters or the parameter 
estimates from different test sets have to lie. We claim that inconsistencies in the estimated parameter values, considering their approximated confidence ellipsoids as well, cannot be explained by data uncertainty but are indicators of model uncertainty.
The proposed method is demonstrated using a component of the 
\mbox{3D Servo Press}, a multi-technology forming  machine that combines spindles with eccentric servo drives.
\end{abstract}

\maketitle

\thispagestyle{empty}

\section{Introduction}

In science, technology and economics mathematical models are commonly used to describe physical phenomena, to solve design 
problems and to manage production processes. The employment of these models frequently entails uncertainty. It has been observed 
that the dominant uncertainties arise from our lack of knowledge about system parameters and from deficiencies in the modeling 
itself \cite{oden2010computer}. We consider \emph{models} to be mathematical constructs which describe the relations 
between inputs, internal variables and outputs. All present knowledge about the technical system or phenomenon of interest is represented 
by such a model. Correspondingly, we mean by \emph{model uncertainty} that some of these functional relations 
are imperfect, insufficient or simplified in comparison to observed reality. Thus, the present description of the system or phenomenon is 
incomplete in the sense that there are aspects which have been ignored. As a consequence, any simulated process or 
manufactured product that is based upon these models is impaired in its predictive quality or usage. Hence, it is important to 
develop tools and algorithms for the identification, quantification and control of model uncertainty. \par

In order to detect whether a model is inadequate, one has to compare the model output to actual experimental data. It is, 
however, difficult to derive a simple criterion for the model to be accurate, since the measurement data is imperfect and 
subject to uncertainty as well. Generally, data uncertainty arises from irreducible randomness, which is also referred to as aleatoric uncertainty \cite{lemaire2014mechanics,zang2002aerospace}, and from systematic errors in the 
measurement process due to lack of knowledge or ignorance, also known as epistemic uncertainty \cite{Roy2011Uncertainty,vandepitte2011symposium}. 
In the course of model calibration, the model parameters are adjusted such as to make the model output compatible to 
experimental observations. As a consequence, uncertainty is transferred from experimental data to the model parameters. \par

In this paper we propose an algorithm to detect model uncertainty using parameter identification, the optimal design 
of experiments approach and statistical hypothesis testing. Here, we understand \mbox{\emph{parameter identification}} to be 
the process of adjusting model parameters as described above and \emph{optimal design of experiments} to be the best 
choice among experimental setups, e.g., sensor types and positions, such that the uncertainty in the estimated parameters is minimized \cite{alexanderian2016design}. Our methodology is able to distinguish between data uncertainty on the 
one hand and model uncertainty on the other hand. Particularly, we interpret any inconsistency in parameter estimates from 
different measurement series as an indicator that the underlying mathematical model is unable to describe \emph{all} measurement series 
with the \emph{same} set of parameter values. We assume neither an a priori distribution nor a specific form of model uncertainty in 
the mathematical equations. \par

The first step before estimating model parameters is to acquire measurements that capture the behavior of the system well. This 
step can be costly if many physical properties of the system have to be observed in each experiment. In some engineering 
applications, measurements are rather taken from a small-sized prototype than from the expensive product 
which is often unavailable yet. It is therefore desirable to know \emph{beforehand} the optimal sensor positions in view of the actual product 
by considering the experimental results from the prototype. Thus, it is valuable to reduce the number of sensors if this does 
not downgrade the reliability of the identified model parameters. Additionally, removing unreliable sensors may even improve the 
quality of the estimate. 
This can be done using the methodology from optimal design of experiments, i.e., by deciding which sensors are actually best suited for 
gathering data in order to minimize the posterior
variance of the estimated parameters. Using these kinds of sensors and their optimal positions, measurements 
with maximum informational value can be 
obtained. \par

To determine model uncertainty based on measurements obtained from an optimally designed experiment, 
we split the experimental data into a calibration and a validation set. Then we solve
the parameter 
identification problem for the calibration set. 
Furthermore, we compute a confidence ellipsoid for a given confidence level $ 1 - \alpha $, where $ \alpha \in (0,1) $, and if the model is correct, then the solution 
of the  parameter 
identification problem for the validation set should lie within this confidence ellipsoid. If the optimal parameters for the 
validation set 
are outside this confidence ellipsoid then we have an indication of model uncertainty. The splitting of the data and the testing is repeated until the number of desired test scenarios is reached.\par

In the literature, a variety of methods exist to detect, quantify and control model uncertainty. We generally 
distinguish between a non-probabilistic approach where uncertainty is treated rather analytically 
\cite{farajpour2012error,simani2003model,smith2014uncertainty}, a probabilistic Bayesian inference based approach to 
assess the prediction quality of a model 
\cite{farajpour2012error,gu2018scaled,mallapur2018quantification,mallapur2019uncertainty,sankararaman2011model,wang2009bayesian} 
and a probabilistic frequentist perspective \cite{liu2011toward,wong2017frequentist,zhao2017validation}. In this paper 
we adopt a probabilistic \emph{frequentist} point of view to deal with model uncertainty. In the following, we explain 
in more detail the main differences to other methods that are closely related to our approach.\par

Model uncertainty is especially discussed in model-based fault diagnosis of machines. Simani et 
al.~\cite{simani2003model} treat uncertainty in the modeling by bounded error terms in the model equations and thus take a robust optimization point of view. This method 
assumes a priori information on the uncertainty in the mathematical equations. In our approach we do not need any assumptions upon the specific form of uncertainty. \par

Our methodology is similar to the idea of K\"orkel et al.~\cite{koerkel2004}, Bauer et al.~\cite{bauer2000oed} and Galvanin et 
al.~\cite{galvanin2007model} who also combined optimal design of experiments with parameter identification. However, they only 
used this method to reliably find optimal parameter values. Asprey and Macchietto \cite{asprey2000statistical} continued this 
methodology to choose between competing models via maximizing a measure of divergence between model predictions. In our approach 
no such measure is needed, we only employ the parameter estimates and their covariance matrices. Another difference is that we 
also consider higher order derivatives in the computation of the covariance matrix \cite{bard1974nonlinear} which is used to determine the confidence 
ellipsoid. \par

There is extensive literature on Bayesian parameter calibration and validation. However, there seem to be only a few references 
dealing with model uncertainty from a general viewpoint. Lima et al.~\cite{lima2017selection} describe a general method to 
select the best model based on Occam's Plausibility Algorithm \cite{oden2017predictive} and Bayesian calibration. However, we do 
not adopt a Bayesian perspective but we involve design of experiments instead to sharpen the parameter estimates. \par

Staying within this Bayesian framework the same question whether a set of measurements for a given model is adequately described 
by the same set of parameters is 
addressed by Tuomi et al.~\cite{tuomi2011application}. Using a given prior distribution for the 
parameters, they derive an inequality to dismiss the veracity of a model. If the probability for the data to be obtained under 
different parameter sets is significantly higher then the model is rejected. In this work we discuss the same question but from a probabilistic frequentist point of 
view without any assumptions on the prior distribution of the parameters.\par

Another important approach to identify and control model uncertainty was introduced by Kennedy and O'Hagan 
\cite{KennedyOHagan2001}. 
This method is based on the assumption that the true values of the quantities of interest are the sum 
of the model output 
$h(p, q)$, with input $q$ and model parameters $p$, and the model discrepancy term $\delta(\theta, q)$. Thus, the measurements $z$ should satisfy the equation
\begin{align}\label{eq:KennedyOHaganModel}
 z = h(p, q) + \delta(\theta, q) + \varepsilon
\end{align}
with independent observational noise $\varepsilon$. Then parameter identification can be performed for 
\eqref{eq:KennedyOHaganModel} to obtain best 
guesses for both the model parameters $ p $ as well as the parameters $ \theta $ of the model uncertainty $\delta$. Arendt 
et al.~\cite{arendt2012quantification} use this approach for model updating and to distinguish between the effects of model  
calibration and model discrepancy. However, it has been shown by Brynjarsd\'{o}ttir and 
O'Hagan \cite{BrynjarsdottirOHagan2014} that the success of this approach heavily depends on incorporating a priori knowledge of 
the specific form of model uncertainty into the representation of $\delta$, which is often assumed to be a specific type of 
stochastic process,
but is actually not known beforehand. In contrast, our approach does not need any assumptions about the specific form of model uncertainty. \par

One particular case of technical systems with a multitude of uncertain parameters and unknown physical effects that challenge 
the modeling process are forming presses. Forming presses are highly loaded machines, which have kinematic degrees of freedom to 
perform a motion and to apply high magnitude forces on a workpiece. During this motion, the workpiece is then formed into a new shape. 
This can cause a considerable deflection of machine components which is of high technical importance. Therefore, we want to model this deformation accurately. 
In this paper, we consider a mechanical forming machine, the 3D Servo Press, that consists of a linkage mechanism. The 
kinematic chain is determined by multiple mechanical components with a large number of parameters. We approximate this chain by a lumped parameter system to reduce the 
number of parameters. When modeling a machine we typically pursue one of two 
objectives that lead to different lumped parameter models: an accurate elastic behavior at low frequencies or an accurate 
frequency response 
\cite{dresig14schwingungen}. In the case at hand we seek a model that represents an accurate elastic behavior at low excitation 
frequencies. To estimate the stiffness of components with non-uniform cross-sections, a finite element model is a typical 
technique. In a second step, the finite element model is reduced to the lumped parameter model. This model order reduction 
makes the model inaccurate besides a variety of uncertain influencing variables like material properties and inexact geometries.
Hence, for some components 
it is necessary to identify the stiffnesses \emph{after} the assembly of the machine. 
Due to the deflection, a relative movement of the components occurs and as a result friction dissipates a portion of this 
kinetic energy. However, for the modeling of friction on a macroscopic level, multiple phenomenological models exist so far
\cite{Bertotti.2006}. In this work, three different friction models are portrayed as competing to explain the load-displacement curve of the 3D Servo Press. We apply our methodology to identify uncertainty in these models and to select the most accurate of them. \par

The paper is organized as follows. First we introduce the parameter identification problem and its covariance estimation. Based on the 
resulting covariance matrix, we then formulate the problem of optimal experimental design to find optimal sensor positions which lead 
to the smallest variance of the resulting parameter estimates. In Section \ref{sec:DetectingModelUncertainty} we 
describe in more detail how parameter identification, optimal design of experiments and hypothesis testing can be used 
to detect model uncertainty. Afterwards we introduce the working principle and the 
mathematical models of the 3D Servo Press. The application of our proposed method to the models of the 3D Servo Press is done in 
Section \ref{Sec:NumericalResults}, where we also present numerical results. We end the paper by giving some concluding remarks.

\section{The Parameter Identification Problem and its Covariance Estimation}\label{sec:ParamIdent}

In this section we present the parameter identification problem in a similar way as it is done by K\"orkel et al.~\cite{koerkel2004}. We first introduce 
some basic notation and assumptions, formulate the problem and then deduce the covariance matrix as well as the considered 
confidence regions. \par

The mathematical model is given by the state equation 
\begin{equation}\label{eq:state}
  E(y, p, q) = 0,
\end{equation}
where $E\colon\R^{d_{y}}\times\R^{n_p}\times\R^{d_q} \to \R^{d_E}$ is an operator coupling the state vector 
$y \in \R^{d_y}$ and the parameters $p \in \R^{n_p}$ for any input variable $q \in \R^{d_{q}}$.
This state equation may be a discretized form of a partial differential equation with large dimensions $ d_y $ and $ d_E $. We assume that \eqref{eq:state} has a unique solution $y $ for any given $p$ and $q$. In our modeling, the input variables represent external boundary or load forces which are applied to a mechanical system, see Section~\ref{3dservo_section}. Particularly, we have $ n_q $ inputs in a loading-unloading scenario and we write $ q_j \in \R^{d_{q}} $ for one input from such a scenario and $ y_j \in \R^{d_y} $ for the corresponding state.  \par

The model parameters $ p $ are in general not known beforehand. Therefore, we need measurements to obtain appropriate 
estimates. Let $ n_S $ denote the number of allocated sensors for data collection. We define a \emph{measurement series} $ z_i $ to be a set of data points $ z_{ijk} $ acquired for all input variables $ j = 1, \ldots, n_q $ and for all sensors $ k = 1, \ldots, n_S $. We collect $ n_M $ different measurement series in order to improve the information gain and accuracy. We assume that the measurements $ z_{ijk} $ are collected by prepositioned sensors where each sensor $k$ has a 
constant standard deviation $\sigma_k \in \R$ for each input $ q_j $ and in each measurement series $ i $. The 
aim of the parameter identification problem is to find
model parameters $ \overbar{p} \in \R^{n_p}$ that best fit the model output to the measurements $ z \in \R^{n_M \times n_q \times n_S} $ for given 
inputs. \par

In general, it is not possible to measure all of the state components directly. Therefore, we introduce an observation operator 
$(y_j, p, q_j) \mapsto \overbar{h}(y_j, p, q_j) \in \R^{n_S}$ that maps state, parameters and inputs to the actual quantity 
that is 
measured. Since we will later choose an optimal subset of all possible sensors, we introduce binary weights $ \omega \in \setenbrace{0, 1}^{n_S} $ 
such that $ \omega_k = 1 $ if and only if sensor $ k $ is used. \par

We apply the least-squares method to find the optimal parameter values which minimize the discrepancy between given measurements $ z $ and the model output weighted by the standard deviation of each sensor, respectively:
\begin{equation}\label{eq:ausgleichsproblem}
\begin{aligned}
	\min\limits_{(y, p)} & \quad \sum_{k = 1}^{n_S} \sum_{j = 1}^{n_q} \sum_{i = 1}^{n_M} \dfrac{\omega_k}{2} 
	\enbrace{\dfrac{z_{ijk} - \overbar{h}_{k}(y_{j}, p, q_j)}{\sigma_{k}}}^2  \\
  \mathrm{s.t.} & \quad E(y_{j}, p, q_j) = 0, \quad \text{for }  j \in \setenbrace{1, \ldots, n_q}.
\end{aligned}
\end{equation}

\begin{remark}
Alternatively, we can also  assume that each sensor $ k $ has a given standard deviation $\sigma_{ijk}$ in each measurement 
scenario $ i \in \setenbrace{1, \ldots, n_M} $ and for each input $ q_j, $ \linebreak $ j \in \setenbrace{1, \ldots, n_q} $. 
However, to keep notation simple, we assume the working precision $\sigma_{k}$ of each sensor to be constant over all 
measurement series and all inputs.
\end{remark}

For convenience, we rewrite problem 
\eqref{eq:ausgleichsproblem} in vector form of dimension $ n = n_M n_q  n_S $ and eliminate the state equation by 
inserting the unique state solution 
$$ y(p) \coloneqq (y_1(p), y_2(p), \ldots, y_{n_q}(p)) = (y(p,q_1), y(p,q_2), \ldots, y(p,q_{n_q}))$$
into the objective function leading to the optimization problem
\begin{equation}\label{eq:ausgleichsproblem_red}
\begin{aligned}
	\min\limits_{p} \; f(p, z, \Omega) \coloneqq \frac{1}{2} r(p, z)^\top \Omega \, r(p, z)
\end{aligned}
\end{equation}
with the notations
\begin{align}
r(p, z) & \coloneqq\Sigma^{-1}\left(z - h(y(p), p, q)\right) \in \R^{n}, \notag\\
h(y(p),p,q)& \coloneqq \mathrm{rep}\enbrace{\barenbrace{\overbar{h}(y_j(p),p,q_j)}_{j = 1, \ldots, n_q}, n_M} \in \R^{n}, \notag\\
\Omega & \coloneqq\mathrm{Diag} \enbrace{ \mathrm{rep} \enbrace{ \barenbrace{\omega_k}_{k = 1, \ldots, n_S}, n_q n_M}} \in \R^{n \times 
n}, \label{eq:def_Omega}\\
\Sigma & \coloneqq\mathrm{Diag} \enbrace{ \mathrm{rep} \enbrace{\barenbrace{\sigma_k}_{k = 1, \ldots, n_S}, n_q n_M}} \in \R^{n \times n}, \notag
\end{align}
where $ \mathrm{rep}(x, m) $ is the repetition function that produces $ m $ copies of the vector $ x $.
Thus, the vector $h$ is an arrangement of $\overbar{h}(y_j(p),p,q_j)$ for all $ j = 1, \ldots, n_q $ in a row vector copied 
$ n_M $ times, while $ \Omega $ and $ \Sigma $ are diagonal matrices consisting of $ n_q n_M $ copies of 
$\omega_1,\ldots,\omega_{n_S}$ and $\sigma_1,\ldots,\sigma_{n_S}$, respectively. The measurement tensor $ z $ is vectorized 
compliant with $ h $ and for convenience we use the same symbol. \par

Problem \eqref{eq:ausgleichsproblem_red} can be (locally) solved using, e.g., an extended Gauss-Newton method, see Dennis 
et al.~\cite{Dennis1981} for more details. We denote the (local) solution of this optimization problem by $ p(z, \Omega) $ to emphasize 
its dependence on the measurements and on the weights. \par

For the quantification of data uncertainty we assume the measurement errors to be
normally and independently distributed, i.e.,
\[ 
z_{ijk} = z^\star_{ijk} + \varepsilon_{k}, \quad \text{ with } \; 
\varepsilon_{k} \in \ND \left(0, \sigma_{k}^2\right),                  
\]
where $z^\star$ are the true (but unknown) values of the quantities that are measured. 
Since the measurement series $ z_i $ are realizations of the same random variable $ Z $, the estimated parameters 
$ p(Z, \Omega) $ are also random variables. Denote the (unknown) expected value of the distribution of $ p(Z, \Omega) $ by 
$ p^\ast $. We are now interested in how a perturbation of $ Z $ propagates to $ p(Z, \Omega) $. 
Therefore, we linearize the solution operator $ Z \mapsto p(Z, \Omega) $ of the parameter identification problem around some fixed 
$\overbar{z}$, which will be specified later, such that the linearized $ p(Z, \Omega) $
is Gaussian distributed, compare, e.g., Proposition 3.2 in~\cite{Ross2010}. Its covariance matrix is defined by
\begin{equation}\label{eq:def_covariance_matrix}
\begin{aligned}
  C(p^\ast, \Omega) &\coloneqq \E\barenbrace{\Bigl(p(Z, \Omega) - p^\ast \Bigr) \Bigl( p(Z, \Omega) - p^\ast\Bigr)^\top }.
\end{aligned}
\end{equation}
Thus, the approximated confidence ellipsoid for a certain confidence level $ 1 - \alpha $, where $ \alpha \in (0,1) $, of the  
multivariate Gaussian distributed solution of the parameter identification is given by
\begin{equation}\label{ConfEllips}
\begin{aligned}
  G\enbrace{\alpha, p^\ast, C(p^\ast,\Omega)} = \setenbrace{p \in \R^{n_p} : (p - p^\ast)^\top  C(p^\ast,\Omega)^{-1} (p - p^\ast) \leq \gamma^2(\alpha)},
\end{aligned}
\end{equation}
where $\gamma^2(\alpha) \coloneqq \chi_{n_p}^2(1-\alpha)$ is the quantile of the $\chi^2$ distribution with $n_p$ degrees of freedom.
For more details on multivariate Gaussian distributions and confidence ellipsoids, see for example Scheff\'e~\cite{Scheffe1960}. 

To derive an analytical expression of the covariance matrix $C$ in \eqref{eq:def_covariance_matrix}, following Bard~ \cite{bard1974nonlinear}, we use standard methods for
the linearized version of the mapping $ Z \mapsto p(Z, \Omega) $ around some $ \overbar{z} $, such that $ p(\overbar{z}, \Omega) $ is a good approximation of $ p^\ast  $. Denote $ \overbar{p} \coloneqq p(\overbar{z}, \Omega) $ for brevity. Then
\begin{align*}
p(Z, \Omega) \approx p(\overbar{z}, \Omega) + \dpart{z} p(\overbar{z}, \Omega) \cdot (Z - \overbar{z}).
\end{align*}
The sensitivity $ \dpart{z} p(\overbar{z}, \Omega) $ can then be determined using the first order optimality condition for the parameter identification problem \eqref{eq:ausgleichsproblem_red}, i.e, 
\begin{equation}\label{eq:KKT}
\dpart{p} f(\overbar{p}, \overbar{z}, \Omega) = 0.
\end{equation}
In order to use the implicit function theorem, we make the following assumption:
\begin{assumption}\label{annahme_func}
  \hspace{50pt}
  \begin{enumerate}
    \item[(i)] $ f(\overbar{p}, \overbar{z}, \Omega) $ is twice continuously differentiable with respect to $ p $.
    \item[(ii)] $ \partial_{pp}^2 f(\overbar{p}, \overbar{z}, \Omega) $ is invertible.  
  \end{enumerate}
\end{assumption}
\begin{remark}
  Note that Assumption \ref{annahme_func} (i) is implied by the condition that the observation operator $ h $ is twice continuously differentiable with respect to $ p $.
\end{remark}
Using Assumption \ref{annahme_func}, we now can apply the implicit function theorem. Thus, equation~\eqref{eq:KKT} 
implicitly defines a mapping $ Z \mapsto p(Z, \Omega) $ and its sensitivity $ \dpart{z} p(\overbar{z}, \Omega) $ is 
given by
\begin{equation}\label{implicitfunctiontheofp}
  \partial_{pp}^2 f(\overbar{p} , \overbar{z}, \Omega)\, \dpart{z} p(\overbar{z}, \Omega) \, \cdot \delta Z = - \partial_{pz}^2 f(\overbar{p} , \overbar{z}, \Omega) \, \cdot \delta Z
\end{equation}
in any direction $ \delta Z $. More precisely, we have
\begin{align*}
  \partial_p f(\overbar{p} , \overbar{z}, \Omega) = & \ r(\overbar{p} , \overbar{z})^\top \myGewichte \, \partial_p r(\overbar{p} , \overbar{z}), \\
  \ddpart{p}{z} f(\overbar{p} , \overbar{z}, \Omega) = & \ \dpart{p} r(\overbar{p} , \overbar{z})^\top \myGewichte \, \dpart{z}r(\overbar{p} , \overbar{z}) = \dpart{p}r(\overbar{p} , \overbar{z})^\top \Omega \Sigma^{-1} , \\
  \ddpart{p}{p} f(\overbar{p} , \overbar{z}, \Omega) = & \ \dpart{p} r(\overbar{p} , \overbar{z})^\top \myGewichte \, \dpart{p} r(\overbar{p} , \overbar{z}) + \sum_{i=1}^n r_i(\overbar{p} , \overbar{z}) \, \myGewichte_{ii}\, \ddpart{p}{p} r_i(\overbar{p} , \overbar{z})\, .  
\end{align*}
Let us define
\begin{equation*}
 H(\Omega) \coloneqq \ddpart{p}{p} f(\overbar{p} , \overbar{z}, \Omega) = J(\Omega)^\top\myGewichte J(\Omega) + S(\Omega)
\end{equation*}
with $ J(\Omega) \coloneqq \dpart{p} r(\overbar{p} , \overbar{z}) $ and
\begin{align*}   
  S(\Omega) & \coloneqq \sum_{i=1}^n r_i(\overbar{p} , \overbar{z}) \, \myGewichte_{ii}\, \ddpart{p}{p} r_i(\overbar{p} , \overbar{z}),
\end{align*}
where $ J(\Omega) \in \R^{n \times n_p} $ and $ S(\Omega) \in \R^{n_p \times n_p} $. The exact calculation of  $J(\Omega)$ and $ S(\Omega) $ is given in the appendix, which requires the following assumption to allow the usage of the implicit function theorem:
\begin{assumption}\label{annahme_func_State}
\hspace{50pt}
\begin{enumerate}
	\item[(i)] The state equation $E$ is twice continuously differentiable in all arguments.
	\item[(ii)] $\partial_{y} E(y(\overbar{p}), \overbar{p} ,q)$ is invertible.   
\end{enumerate}
\end{assumption} \par

We want to make sure that the principal part $ J(\Omega)^\top\myGewichte J(\Omega) $ stays invertible when changing the values of the weights $ \Omega $.
  
\begin{assumption}\label{Ass:InvertibilityH}
The matrix $ \Omega J(\Omega) $ has full column rank, i.e., $\mathrm{rank}(\myGewichte J(\Omega)) = n_p$.
\end{assumption}

From this assumption we can infer invertibility of $ J(\Omega)^\top\myGewichte J(\Omega)$, compare K\"orkel 
et al.~\cite{koerkel2004} for more details. Notice, that Assumption~\ref{Ass:InvertibilityH} cannot be satisfied if 
$ n_S < n_p $ and $ J(\Omega) $ is independent of the inputs. Since the latter could often be the case we require the experimenter 
to employ at least as many sensors as the 
number of parameters which shall be estimated. This will become an important constraint later 
in the optimal experimental design problem in Section~\ref{opt_design_exp}. \par

From \eqref{implicitfunctiontheofp} we obtain
\begin{equation*}
  \dpart{z} p(\overbar{z}, \Omega) = - H(\Omega)^{-1} J(\Omega)^\top \Omega \Sigma^{-1}.
\end{equation*}
Using the calculations from above, the approximated covariance matrix is given by
\begin{equation}\label{CovarianceMatrix}
\begin{aligned}
  C(\overbar{p},\Omega) & = \E \barenbrace{\dpart{z} p(\overbar{z}, \Omega) \cdot (Z - \overbar{z}) (Z - \overbar{z})^\top  \cdot \dpart{z} p(\overbar{z}, \Omega)^\top }\\
  & = \dpart{z} p(\overbar{z}, \Omega) \cdot \E[\varepsilon \varepsilon^\top] \cdot \dpart{z} p(\overbar{z}, \Omega)^\top = \dpart{z} p(\overbar{z}, \Omega) \ \Sigma^2 \ \dpart{z} p(\overbar{z}, \Omega)^\top \\
  & = H(\Omega)^{-1}  J(\Omega)^\top  \myGewichte \, \myVar^{-1}  \myVar^2  \myVar^{-1} \myGewichte J(\Omega)  
H(\Omega)^{-\top} \\
  & = H(\Omega)^{-1}  J(\Omega)^\top \myGewichte^2 \, J(\Omega)  H(\Omega)^{-\top}.
\end{aligned}
\end{equation}

\section{Optimal Design of Experiments}\label{opt_design_exp}

The optimal design of experiments problem deals with the task of finding an optimal experimental configuration such that the 
reliability of the estimated model parameters is maximized. In the case at hand, this task simplifies to determining optimal
sensor positions. Notice, however, that the reliability also depends on the accuracy of the sensors that are used for the measurements, 
whereby each sensor $k$ has a given constant variance $\sigma_{k}^2$. Often, the measurement error is composed of a variety of causes, e.g., the repetition error and internal approximation errors as specified by the manufacturer. Whereas the experimenter is in charge to keep the repetition error small during the experiment, the internal errors are fixed by manufacturing of each sensor. \par

It is very common to measure the reliability of the parameter estimation by a single-valued design function $ \Psi $, see Bauer 
et al.~\cite{bauer2000oed} and Franceschini and Macchietto \cite{franceschini2008model}. It is obvious that a small covariance 
leads to a high reliability of the parameter estimation. However, it is unclear what a small covariance means in terms of 
matrices. In general, there are different approaches how to choose the $ \Psi $ function. We list the most prominent ones 
according to Fedorov and Leonov \cite{fedorov2013optimal}:
\begin{itemize}
  \item \textit{A}-criterion: the trace of the covariance matrix, $ \Psi_A(C) = \mathrm{trace}(C) $,
  \item \textit{D}-criterion: the determinant of the covariance matrix, $ \Psi_D(C) = \det (C) $,
  \item \textit{E}-criterion: the maximal eigenvalue of the covariance matrix, $ \Psi_E(C) = \lambda_\text{max}(C) $.
\end{itemize}
It seems natural to use the \textit{D}-criterion due to its close connection to the volume of the confidence ellipsoid and its invariance with respect to 
transformations applied to the model parameters. However, this criterion tends to emphasize the most sensitive 
parameter \cite{franceschini2008model}. The \textit{A}-criterion ignores the amount of information on the off-diagonal elements 
of the covariance matrix. This is particularly inefficient when there is a high correlation between parameters. For the 
numerical example in this paper, we choose the \textit{E}-criterion even though \textit{E}-optimality may lead to a tolerable increase 
in volume of the confidence ellipsoid. The \textit{E}-criterion effectively reduces the largest expansion of the confidence ellipsoid. \par

We now formulate the optimal design of experiments problem as follows:
\begin{equation}\label{eq:kovarianzmin}
\begin{aligned}
	\min\limits_{\omega} & \quad \Psi \enbrace{C(\overbar{p},\Omega)} \\
	\mathrm{s.t.} 
	& \quad \Omega = \mathrm{Diag} \enbrace{ \mathrm{rep} \enbrace{ \barenbrace{\omega_k}_{k = 1, \ldots, n_S}, n_q n_M}}, \\
	& \quad \overbar{p} = p(\overbar{z}, \Omega) \text{ solution of } \eqref{eq:ausgleichsproblem_red}, \\
	& \quad g(\omega) \leq 0, \quad \omega \in \setenbrace{0, 1}^{n_S}.
\end{aligned}
\end{equation}
The possibly nonlinear constraint $ g(\omega) \leq 0 $ describes further conditions on $ \omega $, e.g., bounds on the number of used sensors. 
In our case, to fulfill the rank condition in Assumption \ref{Ass:InvertibilityH}, the constraint must contain the inequality $ n_p - \sum_{i = 1}^{n_S} \omega_{k} \leq 0 \, $.\par

The optimal design of experiments problem \eqref{eq:kovarianzmin} is thus a non-convex mixed-integer nonlinear program (MINLP). 
Such problems can be solved via spatial branch-and-bound, see, e.g., Burer and Letchford~\cite{BurerLetchford2012} for an 
overview. \par

Note, however, that for the correctness of the proposed approach, problem \eqref{eq:kovarianzmin} does not necessarily need to 
be solved to optimality. Using a good but suboptimal sensor placement will not lead to any incorrect rejection of a model, since
the variance of the parameter estimates becomes larger and therefore also the confidence ellipsoids increase. Thus, it is also 
possible to solve
\eqref{eq:kovarianzmin}, which is the computationally most expensive step of the proposed approach, by heuristic methods. In our 
numerical example, the number of sensors is very small, so that a heuristic method may indeed provide satisfactory results.

\section{Detecting Model Uncertainty}\label{sec:DetectingModelUncertainty}

In this section, we discuss how optimal design of experiments and parameter identification can be used to detect model uncertainty in a mathematical model $ \mathcal{M} $. To do so, assume 
that all parameters of the model have a true physical meaning and that in case the model is correct, the solution of the parameter identification 
problem is a good approximation of those real, physical values. Then repeated solutions of the parameter identification problem for 
different measurements with differing inputs should, within the boundaries of the model up to uncertainty of the measurements, deliver the same set 
of parameters. On the other hand, if one set of measurements leads to parameters which lie outside a given confidence set of the previous runs, then 
this implies that the model cannot replicate the results of all measurements reliably, i.e., the underlying model is inadequate.

Our approach to detect model uncertainty in a mathematical model $ \mathcal{M} $ is depicted in Algorithm~1. As already explained in the introduction, the first step before identifying model parameters by fitting the model output to a given set 
of measurements is to actually acquire these measurements which can be extremely costly. Furthermore, the quality of the parameter estimation may even be improved by 
removing unreliable sensors. Therefore, we only acquire a minimal amount of measurement series, or use artificial data, which is 
needed for the computation of the optimal design of experiments introduced in the previous section to determine optimal sensor 
positions (line~\texttt{02}). In this case, we solve problem \eqref{eq:kovarianzmin} with a restriction on the 
desired number of used sensors to decide which sensors are actually essential to solve the parameter identification 
problem with minimal variance (line~\texttt{03}). \par

After using the optimal experimental setup $ \omega_\mathrm{opt} $ to acquire data it needs to be verified 
whether the measurement errors are normally distributed (lines~\texttt{04}-\texttt{05}). We use the well known Shapiro-Wilk goodness-of-fit test to do so, see D'Agostino~\cite{dagostino1986goodness}. We only consider experiments that render data with Gaussian measurement errors otherwise we cannot apply our algorithm. \par

\begin{table}[t]    
    \renewcommand{\arraystretch}{1.08}\fontsize{10}{11}\selectfont
    \begin{tabular*}{\textwidth}{@{\extracolsep{\fill} }l} 
      \toprule
      \bfseries Algorithm 1 (Detection of Uncertainty in a Mathematical Model) \\
      \midrule
    \end{tabular*}
    \begin{tabular*}{\textwidth}{rl}
       \textbf{Input:} & Model $ \mathcal{M} $, test level $\mathtt{TOL} $ (e.g. $ 5 \% $), number of test scenarios $ n_\mathrm{tests}. $ \\
      \textbf{Output:} & Does $ \mathcal{M} $ need to be rejected? YES (1) or NO (0). 
    \end{tabular*}
    \begin{tabular*}{\textwidth}{>{\footnotesize\ttfamily}rl}
      01: & Initialize $ i \coloneqq 1 $. \\
      02: & Generate initial data $ z^\mathrm{ini} $ in all feasible sensor locations. \\
      03: & Solve \eqref{eq:kovarianzmin} and obtain optimal $ \omega_\mathrm{opt} $. \\
      04: & Acquire measurements $ z $ with the optimal sensor choice for different inputs. \\
      05: & Check whether measurement errors are Gaussian. If not, go to line \texttt{04} or exit. \\
      06: & Divide $ z $ into a calibration set $ z^\mathrm{cal} $ and a validation set $ z^\mathrm{val} $. \\
      07: & Calculate $ \enbrace{p_\mathrm{cal}, C_\mathrm{cal}} $ using $ z^\mathrm{cal} $ by \eqref{eq:ausgleichsproblem_red} and \eqref{CovarianceMatrix}. Likewise, obtain $ p_\mathrm{val} $ using $ z^\mathrm{val} $. \\
      08: & Determine $ \alpha_\mathrm{min} \in (0,1) $, such that $ p_\mathrm{val} $ lies on the boundary of $ G(\alpha_\mathrm{min}, p_\mathrm{cal}, C_\mathrm{cal}) $. \\
      09: & \textbf{if} $ \alpha_\mathrm{min} \geq \mathtt{TOL}/n_\mathrm{tests} $ \textbf{then} \\
      10: & \hspace{2mm} \textbf{if} $ i < n_\mathrm{tests} $ \textbf{then} \\
      11: & \hspace{4.5mm} $ i \coloneqq i + 1 $. Go to line \texttt{06}. \\
      12: & \hspace{2mm} \textbf{else} \\
      13: & \hspace{4.5mm} \textbf{return} 0. \\
      14: & \hspace{2mm} \textbf{end if} \\
      15: & \textbf{else if} $ \alpha_\mathrm{min} < \mathtt{TOL}/n_\mathrm{tests} $ \textbf{then} \\ 
      16: & \hspace{2mm} \textbf{return} 1. \\
      17: & \textbf{end if} \\
      \bottomrule
    \end{tabular*}
\end{table}   

Assume that a test set $ z $ of measurements is given. Then split the test set into one calibration set
$ z^\mathrm{cal} $ and one validation set $ z^\mathrm{val} $, see line~\texttt{06}. This split can either be done randomly, as in 
a Monte Carlo cross-validation \cite{dubitzky2007fundamentals}, or it can be chosen in a way to test whether a specific physical 
effect is sufficiently modeled. 
For example, the test set could be split according to the magnitude of the inputs to check if the results for both sets 
can be reproduced by the model for the same set of parameters. On the one hand, this approach can help to identify ranges of 
input variables for which the model works better or worse and on the other hand, to detect specific effects which are not yet 
sufficiently implemented in the model. \par

From line \texttt{07} onward, a classical hypotheses test with Bonferroni correction \cite{dunn1961multiple} is conducted. For this, the parameters $ p_\mathrm{cal} $ 
and their covariance $ C_\mathrm{cal} $ are computed from the calibration data set $ z^\mathrm{cal} $ using 
\eqref{eq:ausgleichsproblem_red} and \eqref{CovarianceMatrix}, respectively. Likewise, the parameters $ p_\mathrm{val} $ 
are computed from the validation data set $ z^\mathrm{val} $. Now, the following hypothesis is tested:
\begin{align*}
    \mathrm{HYP}_0 \; & : \; p^\ast = p_\mathrm{cal} \text{ is the true parameter value for all inputs } q_j,\\
    \mathrm{HYP}_1 \; & : \; p^\ast \neq p_\mathrm{cal}.
\end{align*}
The corrected threshold $ \overbar{\mathtt{TOL}} = \mathtt{TOL}/n_\mathrm{tests} $ determines the test level which is used to decide whether the null hypothesis 
$ \mathrm{HYP}_0 $ needs to be rejected. If $ p_\mathrm{val} \notin G(\overbar{\mathtt{TOL}}, 
p_\mathrm{cal}, C_\mathrm{cal}) $ then the rejection occurs. Recall, that
\begin{align*}
\begin{aligned}
  G(\overbar{\mathtt{TOL}}, 
  p_\mathrm{cal}, C_\mathrm{cal}) = \setenbrace{p \in \R^{n_p} : (p - p_\mathrm{cal})^\top  C_\mathrm{cal}^{-1} (p - p_\mathrm{cal}) \leq \chi_{n_p}^2\enbrace{1-\overbar{\mathtt{TOL}}}}.
\end{aligned}
\end{align*}
The outcome of the statistical test can easily be determined by comparing its $ p $-value, 
$ \alpha_\mathrm{min} $, with the threshold $ \overbar{\mathtt{TOL}} $ (line~\texttt{09}). The $ p $-value is the smallest test 
level under which the null hypothesis can only just be rejected. If $ \mathrm{HYP}_0 $ cannot pass the test then we detected 
model uncertainty. Otherwise another test is conducted by returning to line~\texttt{06} until the number of desired test scenarios is reached. \par

The Bonferroni correction accounts for the potential problem of multiple testing since we may perform 
the tests on \emph{dependent} validation sets. Without addressing this issue we should expect 
$ \approx n_\mathrm{tests} \mathtt{TOL} $ hypotheses to be rejected, which necessitates the 
introduction of another (arbitrary) threshold to deduce model uncertainty. The very conservative 
Bonferroni correction controls the familywise error rate (FWER), which is the probability of 
rejecting at least one true null hypothesis. By performing $ n_\mathrm{tests} $ tests with the 
modified test level $ \overbar{\mathtt{TOL}} $ we are able to achieve $ \mathtt{TOL} $ as a 
bound for the FWER, which is equivalent to the error of the first kind in multiple hypothesis 
testing. Since \emph{all} individual test levels are drastically reduced we interpret any rejection 
of a null hypothesis as significant, i.e., then model uncertainty is detected and $ \mathcal{M} $ needs to be rejected. \par

In practical applications it may occur that an inaccurate model passes quite a few tests. Evidently, even an inaccurate model may be useful for a small range of input variables. However, a false model will always 
fail at least one test provided that enough data caused by a variety of inputs is available and that the splitting 
into one calibration and one validation 
test set is done intelligently. To catch the worst case in this splitting maneuver, it may be necessary to consult an expert judgment depending on the 
application to properly exploit the special structure of the technical system.

\section{The 3D Servo Press Model}\label{3dservo_section}

The method for detecting model uncertainty is demonstrated at a technical system, the 3D Servo Press 
\cite{scheitza10konzeption}, a forming machine which transmits the torques and forces of its drives onto a part to be 
formed, e.g., a car body part. Therefore, a forming machine is subject to high magnitudes of external forces during 
its motion which cause its mechanism to deflect. While a rigid body model is accurate during the unloaded state, it 
does not suffice during the forming operation \cite{groche17stiffness}. Especially for the closed-loop control of 
forming machines, an accurate model is crucial as inaccuracies can cause the control to become unstable 
\cite{hoppe19closedloop}. However, the modeling of forming machines requires a high degree of abstraction, since 
elastic bodies are usually reduced to bars and beams in order to keep the model tractable. Furthermore, nonlinear bearing stiffnesses as well as friction 
have to be taken into account.
\par

\begin{figure}[t]
  \centering
  \def\svgwidth{0.66\linewidth}
  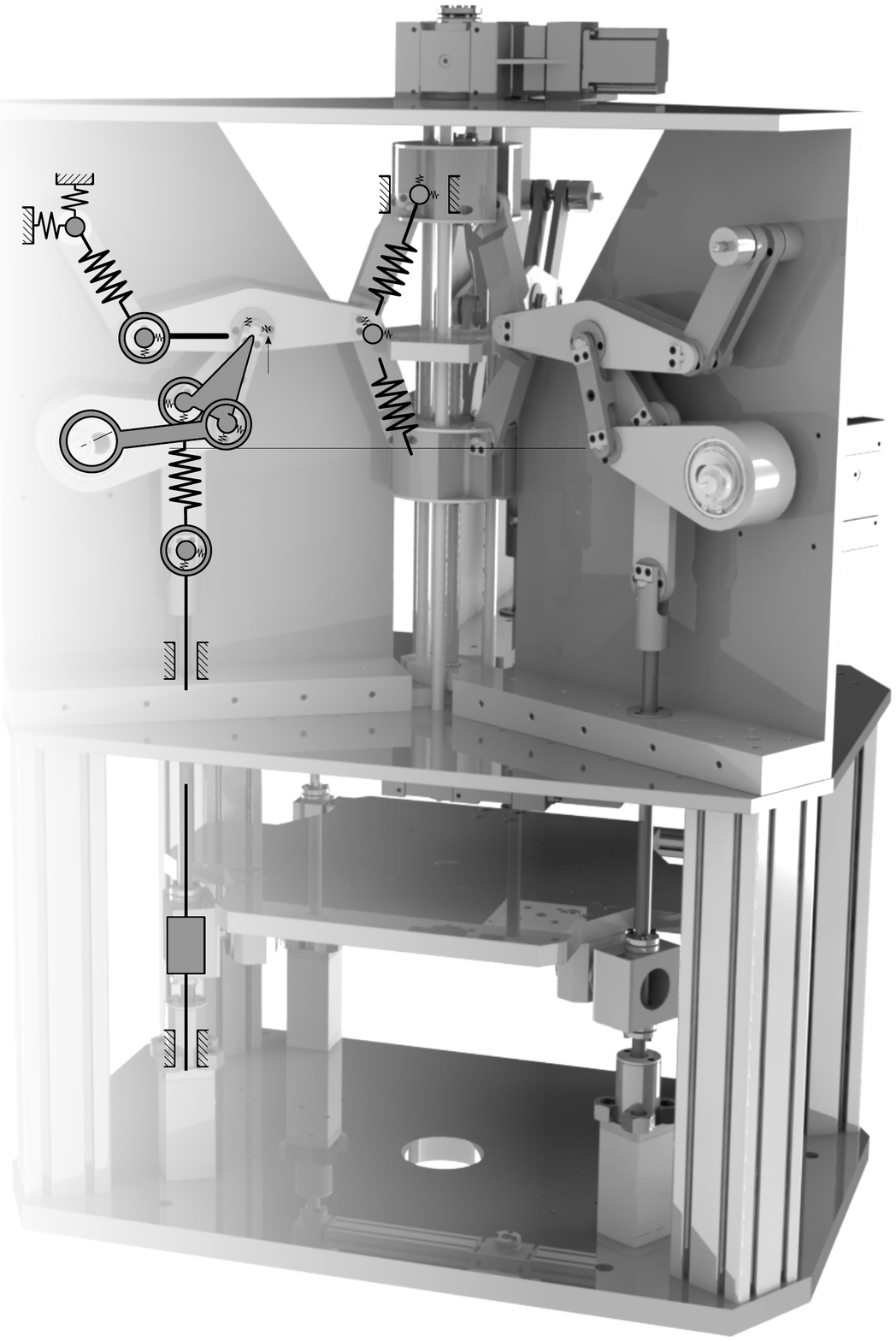
  \caption{Linkage mechanism of the 3D Servo Press.}
  \label{fig:mdl_kin3DSP}
\end{figure}

\figref{fig:mdl_kin3DSP} shows the 3D Servo Press that consists of three identical linkage mechanisms. We use a mechanical substitute model and describe it for one linkage mechanism. A variety of bars and beams are connected via joints that are designed as rotary joints. Each elastic component is represented by a spring or beam and each mass by a gray volume. The eccentric and spindle drives move the three degrees of freedom of one gear unit $\varphi_\mathrm{ecc}$, $y_\mathrm{su}, y_\mathrm{sl}$ that cause all joints in the kinematic chain to perform a desired movement.
The output of the gear unit is point $ D $, which leads down to the ram bearing $ R $ via a linear pressure bar. For the rigid-body model, the position of all points is defined by the angle of the eccentric drive $\varphi_\mathrm{ecc}$ as well as the upper and lower spindle drive position $y_\mathrm{su}, y_\mathrm{sl}$. \par

To model the elastic 3D Servo Press, the coupling links are interpreted as bars and beams, depending on their stress state under load. The bar and beam models are composed of masses and springs. The bearings are modeled as simple spring elements with either linear or non-linear spring characteristics. The equation of motion of the system is determined by the Lagrange equations of the second kind:
\begin{align}
\tdiff{}\left(\ddiff{L}{\dot{y}}\right) - \ddiff{L}{y} = q,
\label{eq:LagrangeEq}
\end{align}
where $L=T-U$ is the Lagrangian consisting of the total kinetic energy $T$ and the total potential energy $U$, $y$ are the system states and $q$ are the non-conservative forces. The non-conservative forces contain all external forces that are applied to the machine, i.e., the torque of the eccentric drive $q_\mathrm{ecc}$, the forces of the upper and 
lower spindles $q_\mathrm{su}$, $q_\mathrm{sl}$ and the reacting process force $q_\mathrm{P}$. In this application we want to evaluate the elastic model and therefore fix the drives positions. Thus, only $q_\mathrm{P}$ is applied and all other non-conservative forces are zero.

Solving the Lagrangian equation requires the potential and kinetic energy as a function of the states. These consist of the stored energy in each elastic and rigid body
\begin{align*}
 T&=\sum_{i=1}^{5} T_{\text{bar},i} + \sum_{i=1}^{1} T_{\text{beam},i} + \sum_{i=1}^{2} T_{\text{body},i},  \\ 
 U&=\sum_{i=1}^{5} U_{\text{bar},i} + \sum_{i=1}^{1} U_{\text{beam},i} + \sum_{i=1}^{10} U_{\text{joint},i},
\end{align*}
whereby the energies of the individual elements are given as follows.

\subsection*{Bar model}
A direct approach to discretizing the bar while maintaining inertia and rigidity is the finite element method. It is based on the partial differential equation of the continuous bar and supplies the mass matrix $M_i$ and the stiffness matrix $K_i$ for an element of mass $m_i$ and stiffness $k_{\text{bar},i}$, which are given by  
\begin{equation*}
	M_i = 
	\begin{bmatrix}
	\frac{1}{2} m_i    & \frac{1}{6} m_i\\
	\frac{1}{6} m_i  & \frac{1}{2} m_i
	\end{bmatrix},
	\quad 
	K_i = 	
	\begin{bmatrix}
	k_{\text{bar},i}    & -k_{\text{bar},i}\\
	-k_{\text{bar},i}  & k_{\text{bar},i}
	\end{bmatrix}.
\end{equation*}
As the actual elements do not have a uniform cross section, the stiffness is determined using a finite element simulation based on the ideal CAD model.

\begin{remark}
  The CAD model and finite element model are based on the detailed knowledge of the elastic modulus and the geometry of the components.
  Due to natural fluctuations in material production, the elastic modulus may vary from part to part. In addition, manufacturing
  limitations only impede geometric accuracy. Therefore, determining the stiffness by an a priori FEM simulation leads to an 
  uncertain estimation of the actual stiffness and requires a parameter identification based on posterior measurements.
\end{remark} 

\begin{figure}[th]
		\small
		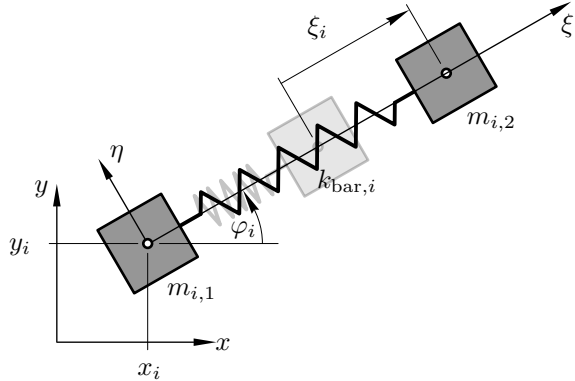
		\caption{Model of a bar consisting of two masses and a spring.}
		\label{fig:ModellStab}
	\end{figure}

The kinetic energy of an individual bar shown in \figref{fig:ModellStab} sums up to
\begin{equation*}
\label{eqn:TModell}
	T_{\text{bar},i} = \frac{1}{2} \left( m_{i,1} v_{i,\text{S}1}^2 + m_{i,2} v_{i,\text{S}2}^2 \right) + \frac{1}{2}  \Theta_i \dot{\varphi}_i^2
\end{equation*}
with the translational velocities of the masses $v_{i,\text{S}j}$, the mass moment of inertia $\Theta_i$ and the corresponding rotational velocity $\dot{\varphi}_i$.
Its potential energy originates from the energy stored in the elasticity and the gravitational potential energy of the masses
\begin{equation*}
U_{\text{bar},i} = \frac{1}{2} k_{\text{bar},i} \xi_i^2 + m_{i,1} g_0 y_{i,1} +m_{i,2} g_0 y_{i,2},
\end{equation*}
where $\xi_i$ is the elongation of the element, $g_0$ is the standard gravity of Earth and $y_{i,j}$ is the relative distance of each mass to the ground. 

\subsection*{Beam model}
\begin{figure}[t]
	\small
		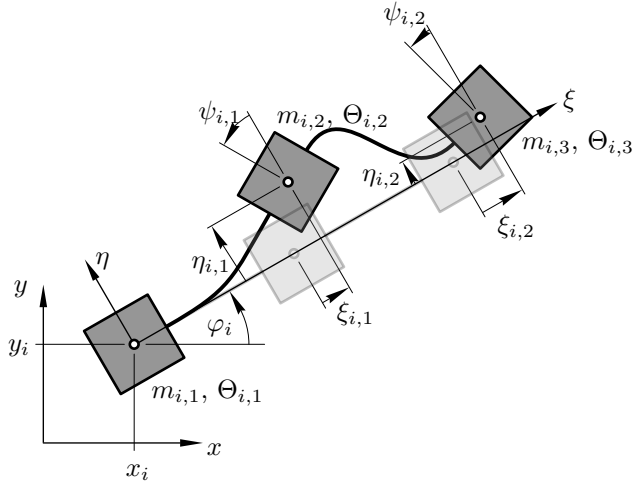
		\caption{Model of a beam consisting of three masses and two springs.}
		\label{fig:ModellBalken}
\end{figure}

All elements that experience bending moments are modeled as beams. This applies especially to the lever, which connects three points instead of two and is marked as a thick gray line in \figref{fig:mdl_kin3DSP}. Like the bar model, the beam model is based on the equations of the finite element method and serves as the basis for modeling the lever under bending load. Since the lever in total features three joints, the model can be seen as two flat beam elements arranged in a row. A lumped mass model is set up in which all elements outside the main diagonal of the mass matrix are neglected.  
The stiffness of each finite element results in a stiffness matrix
\begin{equation*}
\bm{K}_{\text{beam},i,\text{element}} = \begin{bmatrix} k_{i,\alpha} & &  & -k_{i,\alpha} & &  \\
										   &k_{i,\beta} & k_{i,\beta} l_i && -k_{i,\beta} & k_{i,\beta} l_i  \\
										   	&k_{i,\beta} l_i & k_{i,\beta} l_i^2 && -k_{i,\beta} l_i & k_{i,\beta} l_i^2  \\
										   	-k_{i,\alpha} & &  & k_{i,\alpha} &  \\
										   	&-k_{i,\beta} & -k_{i,\beta} l_i & & k_{i,\beta} & -k_{i,\beta} l_i \\
										   	&k_{i,\beta} l_i & k_{i,\beta} l_i^2 & & -k_{i,\beta} l_i & k_{i,\beta} l_i^2  \\
	\end{bmatrix}
\end{equation*}
using the simulated stiffnesses $k_{i,\alpha}$, $k_{i,\beta}$ and the length of the beam $l_i$. Since the lever consists of two finite elements, two $ 6 \times 6 $ element matrices are joined together to form a $ 9 \times 9 $ stiffness matrix according to the finite element method. The result is the stiffness matrix $\bm{K}_{\text{beam},i}$. As shown in \figref{fig:ModellBalken} the total mass of the lever is distributed to the model masses 
\begin{equation*}
	m_{i,1}=\frac{m_i}{4} , \quad m_{i,2}= \frac{m_i}{2} \quad \text{and} \quad m_{i,3} =\frac{m_i}{4}.
\end{equation*}
As the kinetic energy of a beam is equivalent to the kinetic energy of a bar, this results in 
\begin{equation*}
	T_{\text{beam},i} = \frac{1}{2}\sum_{j} m_{i,j} v_{i,\text{S}j}^2 + \frac{1}{2}\sum_{j}   \Theta_{i,j} \dot{\varphi}_{i}^2, 
\end{equation*} 
where $\dot{\varphi}_{i}$ is the rotation of the complete beam.
For the calculation of the potential energy, the sum of the positional energy of the masses and the elastic energy 
\begin{equation*}
	U_{\text{beam},i} =  \bm{y}_{\text{beam},i}^\top \bm{K}_{\text{beam},i} \bm{y}_{\text{beam},i} + \sum_j m_{i,j} g_0 y_{i,j}
\end{equation*}
is calculated where 
\begin{equation*}
	\bm{y}_{\text{beam},i} =  \left[ x_{i,1}, y_{i,1}, \varphi_{i}, \xi_{i,1}, \eta_{i,1}, \psi_{i,1}, \xi_{i,2}, \eta_{i,2}, \psi_{i,2} \right]^\top
\end{equation*}
are the states of the beam.

\subsection*{Bearing model}
The bearings are modeled as spring elements between the joints of the couplers. Since the radial bearing force applied by the bearings is a function of deflection, the deflection must be described with the position coordinates of the bodies. Assuming a constant joint stiffness, the potential energy results in 
\begin{equation*}
U_{\text{joint},i} = \frac{1}{2} k_{\text{joint},i} \; \Delta r_i^{2}
\end{equation*}
with the joint's stiffness $k_{\text{joint},i}$ and its radial deflection $\Delta r_i$.

\subsection*{Friction model}
Friction occurs in all bearings in which a relative movement takes place and will cause a hysteresis in the load-displacement curve. As the relative movements in the joints is small compared to the movement of the pressure bar that connects point $D$ with point $R$ (see Figure \ref{fig:mdl_kin3DSP}), only the bearings guiding this bar are considered. Nevertheless, a variety of model approaches exist for friction. In order to test which approach is the closest to reality in this case, three rate-independent friction models of different complexity are pursued. \par

\begin{enumerate}[label=\arabic*)]
\item Since friction is hard to model, it is often neglected which leads to the model 
\begin{equation*}
  q_{\text{fric}}(t) = 0.
\end{equation*}
\item
The discontinuous Coulomb friction model 
\begin{equation}\label{eq:discontinuousFric}
\begin{aligned}
  q_{\mathrm{fric}}(t) = q_{\mathrm{c}} \; \mathrm{sign} \left( \tdiff{R_x} \right) = q_{\mathrm{c}} \; \mathrm{sign} \left( \tdiff{q_{\mathrm{P}}} \right)
\end{aligned}
\end{equation}
gives a more accurate description of friction in which $q_\text{c}$ is a friction constant. As we can assume that the sign of $\tdiff{R_x}$ is the same as the sign of $\dot{q}_\text{P} = \tdiff{q_\mathrm{P}}$ we can simplify the model to be only discontinuous in the input variables and not in the states. 
\item
As a third model approach, a continuous friction model with rate-independent memory that takes into account past force data is considered \cite{Bertotti.2006}. Here, we take into account the force of the current time step $ t_i $ and the last $ t_{i-1} $:
\begin{align*}
q_\text{fric}(t_i) &= \mu \underbrace{\left( q_{\text{P}}(t_i), q_{\text{P}}(t_{i-1}), q_{\text{P,min}}(t_i), q_{\text{P,max}}(t_i) \right)}_{\bar{u}} ,
\end{align*}
as well as the minimum and maximum force value during loading and unloading cycles
\begin{align*}
q_{\mathrm{P,min}}(t_i) &= \begin{cases}\min(q_{\mathrm{P}}(t_i),q_{\mathrm{P,min}}(t_{i-1})) & \text{if } \dot{q}_\mathrm{P}(t_i)\geq 0\\ q_{\mathrm{P}}(t_i) & \text{if } \dot{q}_\mathrm{P}(t_i)<0 \end{cases}  \\
q_{\mathrm{P,max}}(t_i) &= \begin{cases}q_{\mathrm{P}}(t_i) & \text{if } \dot{q}_\mathrm{P}(t_i) \geq 0\\ \min(q_{\mathrm{P}}(t_i),q_{\mathrm{P,max}}(t_{i-1})) & \text{if } \dot{q}_\mathrm{P}(t_i)<0 \end{cases}
\end{align*}
that are internal variables and reduce the complexity of memorizing a large number of time steps. Based on the Preisach model \cite{Preisach.1935} which is a discontinuous hysteresis model, we used an adapted continuous model which is comparable to a neural network topology \cite{Mayergoyz.2003}. \figref{fig:PreisachMdl} shows the topology of the used model where $\rho_i = \arctan(\bar{u})$.
\begin{figure}
	\centering
		\includegraphics[width=1.0\textwidth, trim=3cm 8cm 3cm 3cm]{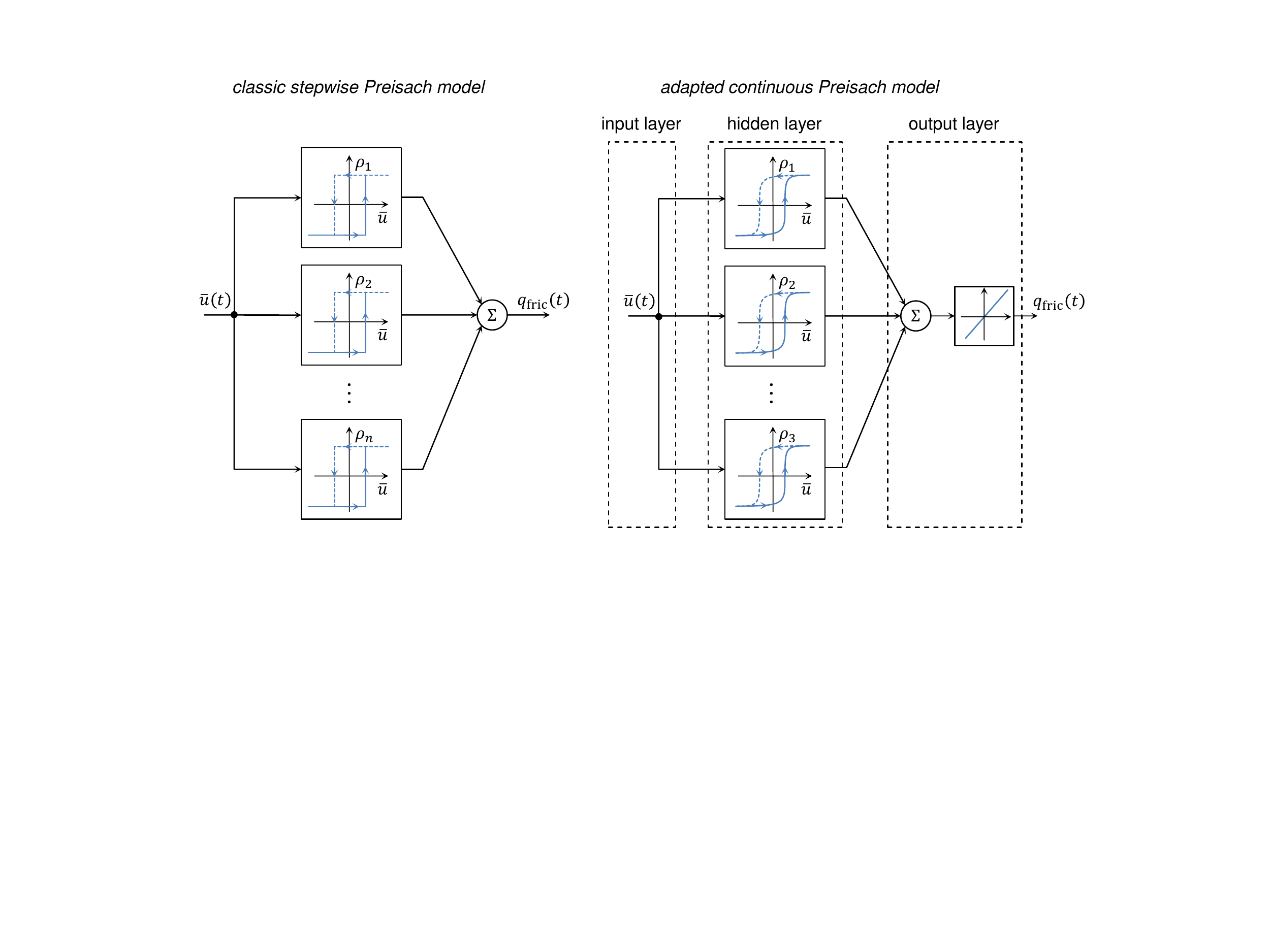}
	\caption{Model topology of the classical discontinuous (left) and the adapted Preisach model (right).}
	\label{fig:PreisachMdl}
\end{figure}
To train the model, we have to determine the friction force which is the difference of the actual measured process force and the estimated force by the inverse model.
The inverse model describes the required force under a measured displacement $z$ and contains the estimated stiffness parameters that have been determined without any friction model in a first step. Applying this to measurements of a loading cycle, the full hysteresis can be identified and used to train the friction model.
\end{enumerate}

\subsection*{Synthesis of the press model}
The press model consists of 2 rigid bodies, 5 bars, 1 beam, 10 joints and the elasticity of the press frame which represents support points to the environment. This results in a 34-dimensional state vector $y$.

Equation \eqref{eq:LagrangeEq} can now be written as
\begin{equation*}
f_\text{kin}(y, \dot{y}, \ddot{y}) + f_\text{pot}(y) = q(t),
\end{equation*}
with the contribution of the kinetic energy $f_\text{kin} \left( \bm{y}, \dot{\bm{y}},t \right)$ and of the potential energy $f_\text{pot} \left( \bm{y},t \right)$ and the excitation forces
\begin{equation*}
q(t) = q_\text{P}(t) - q_\text{fric}(t).
\end{equation*}
In this case we are interested in the quasi-static model to identify uncertain stiffness parameters of two bars $k_{\text{bar},7}$ and $k_{\text{bar},5}$, in the following denoted as $k_7$ and $k_5$ as shown in \figref{fig:mdl_kin3DSP}. Thus, all derivatives of $y$ are set to zero such that
\begin{equation*}
f_\text{kin}(y, \dot{y}=0, \ddot{y}=0) + f_\text{pot}(y) = q(t)
\end{equation*}
where the function $f_\text{kin}$ contains the parameters $k_7$ and $k_5$.

To identify the model parameters, a process force $q_\mathrm{P}$ is applied using an external pneumatic force source.

%

\section{Numerical Results for the 3D Servo Press}\label{Sec:NumericalResults}

We implemented the described procedure to detect model uncertainty using MATLAB R2017b with the included \texttt{lsqnonlin} solver for the parameter 
identification problems and applied it to the gear mechanism model of the 3D Servo Press. \par

We use measurements for $29$ different process forces (these are the input variables), whereby the first $15$ forces describe loading and the last $14$
describe unloading of the 3D Servo Press. For each process force we measure the vertical displacements in point $D$, the 
horizontal displacements in point $F$ and the vertical displacements in point $B_0$ when applying a vertical process load 
$q_\mathrm{P}$ on the press, see \figref{fig:mdl_kin3DSP}. The displacements are measured in $ \upmu \mathrm{m} $ and the forces in $ \mathrm{N} $. \par

In this particular application we do not distinguish between initial data and actual measurements. Thus, line \texttt{04} in Algorithm 1 is omitted. Each measurement is performed $ n_M = 6 $ times \emph{on the prototype} of the 3D Servo Press although with slightly differing forces 
due to variations in the pneumatic pressure when applying the force. Since we know the setpoint values for the applied forces $ q^\mathrm{d}_j $ for all $ j = 1, \ldots, 29 $ we linearly interpolate the measurements $z\in\R^{6 \times 29 \times 3}$ such as to make them comparable for each force $ q^\mathrm{d}_j $, respectively. More specifically, we apply the correction
\begin{align*}
  z_{ijk} = \dfrac{q^\mathrm{d}_j}{q_j}\cdot z_{ijk}
\end{align*}
for all $ i = 1, \ldots, 6 $, $ j = 1, \ldots, 29 $ and $ k = 1, \ldots, 3 $. We work from now with these corrected measurements. \par

In a first step, we analyze the experimental data. In our modeling we assumed that the measurements are normally distributed. Since the true values $ z^\ast $ of the quantities that are measured are unknown to us, we check whether the measurement errors $ \varepsilon $ are normally distributed with zero mean instead. In order to verify this assumption, we perform a Shapiro-Wilk goodness-of-fit-test \cite{dagostino1986goodness} applied to the measurement errors
\begin{align*}
\tilde{z}_k \coloneqq
  \begin{pmatrix}
   z_{2 k j}-z_{1 k j} \\ 
   z_{4 k j}-z_{3 k j} \\ 
   z_{6 k j}-z_{5 k j}
  \end{pmatrix}_{j = 1, \ldots, 29}.
\end{align*}
for each sensor $ k = 1, \ldots, 3 $ with test level $ \alpha = 5 \% $. Evidently, $ z_{1 k j}, z_{2 k j}, \ldots, z_{6 k j} $ are independent and identically distributed with the same mean and the same standard deviation. Hence, the rows in $ \tilde{z}_k $ are independent and identically distributed with mean zero. The hypothesis that each $ \tilde{z}_k $ is normally distributed with mean zero and variance estimated from $ \tilde{z}_k $ is now tested and the results are shown in Table \ref{HypMeasData}. We observe that the hypothesis cannot be rejected with an error of the first kind below $ 5 \% $ for all sensors, respectively.

\begin{table}[tb]
\caption{Analysis of the measurement data.}
\centering 
\begin{tabular*}{0.66\textwidth}{@{\extracolsep{\fill} }lll@{}}
\toprule
Sensor &  $ p $-value (in \%) & Sigma \\
\midrule
1 & 60.11 & \num{5.5147e-06} \\
2 & 79.64 & \num{3.3108e-06} \\
3 & 60.26 & \num{1.4974e-06} \\
\bottomrule
\end{tabular*} 
\label{HypMeasData}
\end{table} 

Having experimental data available, the aim is to reduce the costs for obtaining new measurements in view of 
future experiments \emph{on the real press}, i.e., we want to reduce the number of involved sensors. The parameters to be estimated, $k_5$ and 
$k_{7}$, describe the axial stiffness of elastic components of the 3D Servo Press, see Section \ref{3dservo_section}. Since the 
number of involved sensors must be greater or equal to the number of estimated parameters, compare Assumption 
\ref{Ass:InvertibilityH} and the comments below this assumption, we want to choose two of the three sensors for which the design criterion of the covariance matrix of the estimated parameters becomes minimal. For comparison, we compute all design criteria that are mentioned in Section \ref{opt_design_exp} for the model $ \mathcal{M}_3 $. The results are shown in Table~\ref{tab:res_DOE}. 

\begin{table}[b]
\caption{Outcome for the optimal design of experiments problem for the model $ \mathcal{M}_3 $.}
\centering 
\begin{tabular*}{\textwidth}{@{\extracolsep{\fill}}llll@{}}
\toprule
Sensor combination &  $ \Psi_A(C) $ & $ \Psi_D(C) $ & $ \Psi_E(C) $ \\
\midrule
111 (initial) & \num{4.9592e+09} & \num{1.1682e+16} & \num{4.9568e+09} \\
101 & \num{1.1180e+29} & \num{7.1838e+35} & \num{1.1180e+29} \\
011 & \num{6.2584e+09} & \num{1.4848e+16} & \num{6.2561e+09} \\
110 & \num{3.5140e+09} & \num{2.7566e+16} & \num{3.5062e+09} \\
\bottomrule
\end{tabular*} 
\label{tab:res_DOE}
\end{table} 

We observe that omitting the second sensor increases all design criteria by a factor of $ \approx\! 10^{+20} $ compared to the initial 
sensor configuration, which is an indication that the covariance matrix became close to singular. A removal of the first sensor, 
though, increases the maximal eigenvalue \textit{and} the volume of the confidence ellipsoid slightly. However, omitting the 
last sensor, i.e., measuring the vertical displacements in point $B_0$, leads to the smallest maximal eigenvalue. We choose the \textit{E}-criterion as 
design criterion for reasons explained in Section \ref{opt_design_exp}. Thus, we proceed with the optimal sensor combination 
$ 110 $, i.e., we choose to measure the vertical displacements in point $D$ and the horizontal displacements in point $F$. We come to the same conclusion after investigating the results for the models $ \mathcal{M}_1 $ and $ \mathcal{M}_2 $. \par

\begin{figure}[tb]
\flushleft
\begin{subfigure}[t]{0.36\textwidth}
%
%
%
\begin{tikzpicture}

\begin{axis}[%
width=\textwidth,
height=0.75\textwidth,
scale only axis,
ylabel={displacement of joint $F_x$ in $\upmu$m},
xlabel={force $q_\mathrm{P}$ in N},
legend style={at={(0.01,0.99)},anchor=north west,legend cell align=left, align=left, draw=white!15!black},
xmax =1600,
ymax = 1600,
grid = none,
label style={font=\scriptsize},
tick label style={font=\tiny},
axis x line = bottom,
axis y line = left,
tick align={outside}
]
\addplot [color=black!20!white, mark=square, mark size=1.2pt]
  table[y index=1, x index=0, col sep=comma] {Messungen_Fx_neu.csv};
\addplot [color=black!30!white, mark=diamond, mark size=1.2pt]
  table[y index=2, x index=0, col sep=comma] {Messungen_Fx_neu.csv};
\addplot [color=black!50!white, mark=triangle, mark size=1.2pt]
  table[y index=3, x index=0, col sep=comma] {Messungen_Fx_neu.csv};
\addplot [color=black!70!white, mark=triangle*, mark size=1.2pt]
  table[y index=4, x index=0, col sep=comma] {Messungen_Fx_neu.csv};
\addplot [color=black!80!white, mark=otimes, mark size=1.2pt]
  table[y index=5, x index=0, col sep=comma] {Messungen_Fx_neu.csv};
\addplot [color=black!90!white, mark=star, mark size=1.2pt]
  table[y index=6, x index=0, col sep=comma] {Messungen_Fx_neu.csv};

\end{axis}

\end{tikzpicture}%
\end{subfigure}
\hspace{23mm}
\begin{subfigure}[t]{0.36\textwidth}
\begin{tikzpicture}

\begin{axis}[%
width=\textwidth,
height=0.75\textwidth,
scale only axis,
ylabel={model output $ \mathcal{M}_1 $ in $\upmu$m},
xlabel={force $q_\mathrm{P}$ in N},
legend style={at={(0.01,0.99)},anchor=north west,legend cell align=left, align=left, draw=white!15!black},
xmax =1600,
ymax = 1600,
grid = none,
label style={font=\scriptsize},
tick label style={font=\tiny},
axis x line = bottom,
axis y line = left,
tick align={outside}
]
\addplot [thick, color=black, mark=square, mark size=0pt]
  table[y index=1, x index=0, col sep=comma] {Modeloutput_M1_Fx_neu.csv};
\addplot [color=black!30!white, mark=triangle, mark size=1.2pt]
  table[y index=2, x index=0, col sep=comma] {Messungen_Fx_neu.csv};

\end{axis}

\end{tikzpicture}%
\end{subfigure}
\par\bigskip
\begin{subfigure}[t]{0.36\textwidth}
\begin{tikzpicture}

\begin{axis}[%
width=\textwidth,
height=0.75\textwidth,
scale only axis,
ylabel={model output $ \mathcal{M}_2 $ in $\upmu$m},
xlabel={force $q_\mathrm{P}$ in N},
legend style={at={(0.01,0.99)},anchor=north west,legend cell align=left, align=left, draw=white!15!black},
xmax =1600,
ymax = 1600,
grid = none,
label style={font=\scriptsize},
tick label style={font=\tiny},
axis x line = bottom,
axis y line = left,
tick align={outside}
]
\addplot [thick, color=black, mark=square, mark size=0pt]
  table[y index=1, x index=0, col sep=comma] {Modeloutput_M2_Fx_neu.csv};
\addplot [color=black!30!white, mark=triangle, mark size=1.2pt]
  table[y index=2, x index=0, col sep=comma] {Messungen_Fx_neu.csv};

\end{axis}

\end{tikzpicture}%
\end{subfigure}
\hspace{23mm}
\begin{subfigure}[t]{0.36\textwidth}
\begin{tikzpicture}

\begin{axis}[%
width=\textwidth,
height=0.75\textwidth,
scale only axis,
ylabel={model output $ \mathcal{M}_3 $ in $\upmu$m},
xlabel={force $q_\mathrm{P}$ in N},
legend style={at={(0.01,0.99)},anchor=north west,legend cell align=left, align=left, draw=white!15!black},
xmax =1600,
ymax = 1600,
grid = none,
label style={font=\scriptsize},
tick label style={font=\tiny},
axis x line = bottom,
axis y line = left,
tick align={outside}  
]
\addplot [thick,color=black, mark=square, mark size=0pt]
  table[y index=1, x index=0, col sep=comma] {Modeloutput_M3_Fx_neu.csv};
\addplot [color=black!30!white, mark=triangle, mark size=1.2pt]
  table[y index=2, x index=0, col sep=comma] {Messungen_Fx_neu.csv};

\end{axis}

\end{tikzpicture}%
\end{subfigure}
\caption{Repeated measurements of the force-displacement curve of the linkage mechanism and comparison with the output of the models $ \mathcal{M}_1, \; \mathcal{M}_2 $ and $ \mathcal{M}_3 $.}
\label{fig:meas3DSP}
\end{figure}
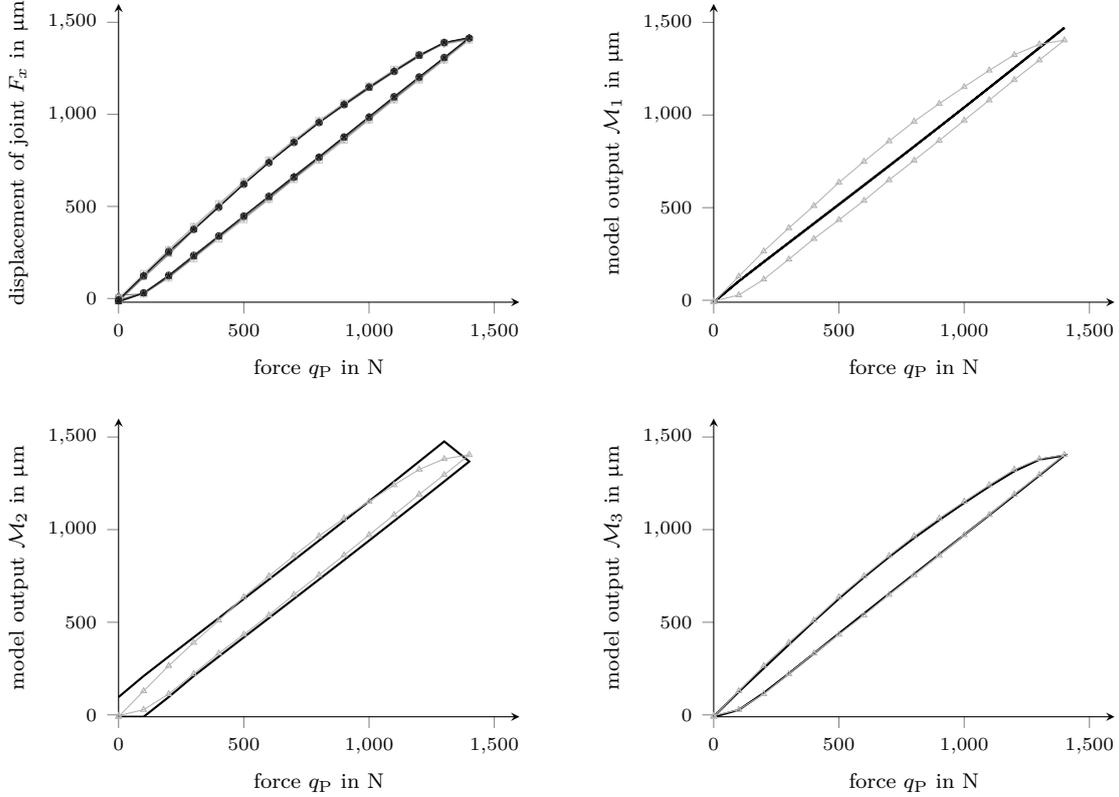

Next, we want to test whether our algorithm recognizes the best out of three different models used to 
describe the data. Therefore, we recall the following friction models from Section~\ref{3dservo_section}:
\begin{align*}
  \mathcal{M}_1 \; & : \; \text{ simple linear model without hysteresis recognition}, \\
  \mathcal{M}_2 \; & : \; \text{ Coulomb's friction model for hysteresis}, \\
  \mathcal{M}_3 \; & : \; \text{ friction behavior learned by a neural network}.
\end{align*}
Figure \ref{fig:meas3DSP} shows the different behavior of these models plotted together with the data.

For the assembly of $ \mathcal{M}_3 $ we need actual measurements to train the neural network as described in Section~\ref{3dservo_section}. For this purpose we employ four data 
series. The remaining two measurement series will be used for the application of our 
algorithm to detect model uncertainty in the press. In order to make the following test strategy fair, we only use these two measurement 
series for \emph{all} models alike since the more data series are involved, the harder it is for a model to reproduce them 
all. \par

The standard deviation of the sensors is crucial for the size of the confidence ellipsoid of the parameter estimates. We fix 
these values to be the weighted sum of the standard deviation of the repeated measurement process, see Table~\ref{HypMeasData}, and other internal errors as specified by the manufacturer of the sensor. Thus, we take the 
values
\begin{align*}
  \sigma_1 & = \sqrt{\left(\num{5.5147e-06}\right)^2 + \left(\num{1.4142e-05}\right)^2} \approx \num{1.518e-05},\\
  \sigma_2 & = \sqrt{\left(\num{3.3108e-06}\right)^2 + \left(\num{3.6055e-06}\right)^2} \approx \num{4.895e-06}, \\
  \sigma_3 & = \sqrt{\left(\num{1.4974e-06}\right)^2 + \left(\num{3.6055e-06}\right)^2} \approx \num{3.904e-06}.   
\end{align*}
In order to investigate the validity of the models $ \mathcal{M}_1, \mathcal{M}_2 $ and $ \mathcal{M}_3 $, we generate 
calibration and validation sets of almost equal size whereby we omit the first $ q^\mathrm{d}_1 = 0 $ 
and last $ q^\mathrm{d}_{29} = 0 $ ``applied'' force because they are referring to the unloaded 
press. We first split 
the test set consisting of the two measurement series into one loading $\mathcal{S}^l$ and one unloading $\mathcal{S}^u$ test set and 
consider each set separately. Thus, 
for the loading set $\mathcal{S}^l$, we again split the test set into one calibration $ \mathcal{S}^{l_1}_c $ and one validation $ \mathcal{S}^{l_2}_v $ test set. We do the same for the unloading case.
Next, we test loading versus unloading and again split the set into one calibration $ \mathcal{S}^{l}_c $  and one validation $ \mathcal{S}^{u}_v $ test set. Lastly, we test loading together with unloading and split the set into one calibration $ \mathcal{S}^{lu}_c $ and one validation $ \mathcal{S}^{lu}_v $ test set, compare Table \ref{Testsets}. The 
splitting is done manually and in this particular way in order to catch the worst case in the coming hypothesis test, 
which we expect to be the case for loading vs. unloading.

\begin{table}[t]
\caption{Summary of the calibration and validation test sets for the different cases.}
\centering 
\begin{tabular*}{\textwidth}{@{\extracolsep{\fill}}lll@{}}
\toprule
Force progression & Calibration & Validation\\
\midrule
Loading & $ \mathcal{S}^{l_1}_c = \{q^\mathrm{d}_2, q^\mathrm{d}_4, \ldots, q^\mathrm{d}_{14} \}$
& $ \mathcal{S}^{l_2}_v = \{ q^\mathrm{d}_3, q^\mathrm{d}_5, \ldots, q^\mathrm{d}_{15} \} $ \\
Unloading & $ \mathcal{S}^{u_1}_c = \{ q^\mathrm{d}_{15}, q^\mathrm{d}_{17}, \ldots, q^\mathrm{d}_{27} \}$ & 
$ \mathcal{S}^{u_2}_v = \{ q^\mathrm{d}_{16}, q^\mathrm{d}_{18}, \ldots, q^\mathrm{d}_{28} \} $ \\
Loading vs.\ unloading & $ \mathcal{S}^{l}_c = \{ q^\mathrm{d}_2, \ldots, q^\mathrm{d}_{14} \}$ & 
$  \mathcal{S}^{u}_v = \{ q^\mathrm{d}_{15}, \ldots, q^\mathrm{d}_{28} \}$ \\
Loading and unloading & $\mathcal{S}^{lu}_c = \{ q^\mathrm{d}_3, q^\mathrm{d}_5, \ldots, q^\mathrm{d}_{27} \}$ & 
$ \mathcal{S}^{lu}_v = \{ q^\mathrm{d}_2, q^\mathrm{d}_4, \ldots, q^\mathrm{d}_{28} \}$ \\
\bottomrule 
\end{tabular*} 
\label{Testsets}
\end{table} 

\begin{table}[b]
\caption{Test results for the 3D Servo Press models $ \mathcal{M}_1, \mathcal{M}_2 $ and $ \mathcal{M}_3 $.}
\centering 
\begin{tabular*}{\textwidth}{@{\extracolsep{\fill}}llrrr@{}}
      \toprule
       Calibration & Validation & $ \alpha_\mathrm{min} $ (in \%)  & $ \alpha_\mathrm{min} $ (in \%)  & 
       $ \alpha_\mathrm{min} $ (in \%) \\
       & & for $ \mathcal{M}_1 $ & for $ \mathcal{M}_2 $ & for $ \mathcal{M}_3 $ \\
      \midrule
      $ \mathcal{S}^{l_1}_c $ & $ \mathcal{S}^{l_2}_v $ & 0.02 & 78.78 & 92.99 \\
      $ \mathcal{S}^{u_1}_c $ & $ \mathcal{S}^{u_2}_v $ & $ \ll $ 0.01 & 23.33 & 66.06 \\
      $ \mathcal{S}^{l}_c $ & $ \mathcal{S}^{u}_v $ & $ \ll $ 0.01 & $ \ll $ 0.01 & 24.59 \\
      $ \mathcal{S}^{lu}_c $ & $ \mathcal{S}^{lu}_v $ & 0.81 & $ \ll $ 0.01 & 93.45 \\
      \bottomrule 
\end{tabular*}
\label{ResultsAll}
\end{table}

For each of the three models and for each of the $ n_\mathrm{tests} = 4 $ test scenarios we perform the hypothesis test as described in 
Algorithm 1 starting from line \texttt{08}. Table \ref{ResultsAll} lists the results. The last three columns show the minimal 
test level, i.e., the $ p $-value, such that the null hypothesis can only just be rejected. We choose the common $ \mathtt{TOL} = 5 \% $ bound for the FWER and apply the Bonferroni correction which reduces the individual test level to $ \mathtt{TOL} / n_\mathrm{tests} = 1.25 \% $. Comparing the values for $ \alpha_\mathrm{min} $, we clearly see that the model 
$ \mathcal{M}_1 $, which does not account for hysteresis, is rejected for all test scenarios. Thus, the data cannot be described by this simple linear model. We demand $ \mathcal{M}_1 $ to be updated 
such as to correctly represent hysteresis. This is done in a first attempt by the Coulomb friction model, see equation 
\eqref{eq:discontinuousFric}. We thus perform our algorithm on $ \mathcal{M}_2 $. While this model seems to be able to describe 
loading and unloading separately, it fails to describe both scenarios with the same set of parameters. Since hysteresis is a 
continuous effect, the discontinuous Coulomb friction model still fails to reproduce the fine nuances of 
the experimental data. Our proposed method is able to detect this deficiency in the third and fourth test scenario, where the model is 
clearly rejected since the $ \alpha_\mathrm{min} $ is very small. Hence, a neural network strategy has been employed to further improve the model output 
as mentioned in Section \ref{3dservo_section}. The last column of Table \ref{ResultsAll} shows that model $ \mathcal{M}_3 $ is 
well-suited to explain the hysteresis phenomenon. \par

To sum up, we have seen that the algorithm is able to detect model uncertainty and by suitable choice of the calibration and 
validation test sets, it can even help to identify (neglected) aspects of the 3D Servo Press model 
that need to be improved. Of course, the modeling errors can also be seen in 
Figure~\ref{fig:meas3DSP} directly. Our algorithm, though, provides an automatized way to decide if a 
model needs to be improved regardless of the dimension of the model's output.

\section{Conclusion}

In this paper we have seen how model uncertainty can be identified by combining the optimal design of experiments approach with  parameter identification and statistical testing.
Optimal design of experiments can be used to choose sensors which allow for parameter estimates with minimal variance.
Using the covariance matrix we can then compute confidence ellipsoids which should include the parameter estimates with
high probability. If some other test set leads to a solution of the parameter identification outside such a confidence ellipsoid then we can conclude
with a small error of the first kind that not all measurements can be explained by the same model with the same set of parameters. We then introduced
the 3D Servo Press as an application and demonstrated our approach on mathematical models of the press. This allowed us to show 
that two simple press models are not valid, since specific effects like hysteresis are not sufficiently modeled. A sophisticated 
mirroring of the hysteresis effect, though, led to a mathematical model that is well-suited to explain the data and thus to make predictions for future experiments.  \par

It would be interesting to further test our method with models that depend on more than two parameters and to have a 
larger number of possible sensor locations available. Furthermore, instead of only choosing sensors once in the beginning, it 
is also possible to re-solve the optimal experimental design problem using the parameters identified through some first experiments to iteratively strengthen the quality of the parameter estimates,
in a similar way as proposed by K\"orkel et al.~\cite{koerkel2004}.

\section{Acknowledgement}

This research was funded by the German Research Foundation (DFG) -- project number 57157498 -- CRC 805 within the subprojects A3, A4 and B2. The authors would like to thank the DFG for funding. 

\bibliographystyle{abbrv}
\bibliography{main}

\section*{Appendix}

To compute $ S(\Omega) $, we need to derive the first and second derivatives of $ r_i(\overbar{p}, \overbar{z}) $ with respect to $ p $.
\begin{equation}\label{eq:partial_r}
\begin{aligned}
   \dpart{p} r_i(\overbar{p}, \overbar{z}) = -\frac{1}{\Sigma_{ii}} & \Big[ \dpart{y} h_i(y(\overbar{p}), \overbar{p} ,q) y'(\overbar{p}) 
   +  \dpart{p} h_i(y(\overbar{p}), \overbar{p} ,q) \Big], \\
  \ddpart{p}{p} r_i(\overbar{p}, \overbar{z}) = - \frac{1}{\Sigma_{ii}} & \Big[ y'(\overbar{p})^\top \ddpart{y}{y} h_i(y(\overbar{p}), \overbar{p} ,q) y'(\overbar{p})
  + 2 \ddpart{y}{p} h_i(y(\overbar{p}), \overbar{p} ,q) y'(\overbar{p}) \\
  & + \dpart{y}h_i(y(\overbar{p}), \overbar{p} ,q)^\top y''(\overbar{p}) 
  + \ddpart{p}{p} h_i(y(\overbar{p}), \overbar{p} ,q) \Big] .
\end{aligned}
\end{equation}
To determine the terms $y'(\overbar{p}) $ and $y''(\overbar{p}) $, we again apply the implicit function theorem, using 
Assumption~\ref{annahme_func_State}, yielding
\begin{equation}\label{firstDerStateEq}
y'(\overbar{p})=-\left(\dpart{y} E(y(\overbar{p}), \overbar{p} ,q)\right)^{-1}\dpart{p} E(y(\overbar{p}), \overbar{p}, q).
\end{equation}
Moreover, the second directional derivatives of $y$ with respect to $ p $ in directions $h_1$ and $h_2$ is given by
\begin{equation}\label{secondDerStateEq}
\begin{aligned}
y''(\overbar{p})(h_1;h_2)
 = & -\left(\dpart{y} E(y(\overbar{p}), \overbar{p} ,q)\right)^{-1}\Big[\ddpart{y}{y} E(y(\overbar{p}), \overbar{p} ,q)(y'(\overbar{p})h_1;y'(\overbar{p})h_2)  \\ & + 2\ddpart{y}{p} E(y(\overbar{p}), \overbar{p} 
,q)(y'(\overbar{p})h_1;h_2) +\ddpart{p}{p} E(y(\overbar{p}), \overbar{p} ,q)(h_1;h_2)\Big].
\end{aligned}
\end{equation}
The exact characterization of the vector-tensor and matrix-tensor products in equation \eqref{eq:partial_r} and \eqref{secondDerStateEq} above is given by
\begin{align*}
\dpart{y}h_i^\top y''(\overbar{p}) = 
- & \left[ \sum\limits_{\ell, m = 1}^{n_p} \sum\limits_{k = 1}^{n_y} \enbrace{\sum\limits_{i, j = 1}^{n_y} \bigl(\dpart{y}h_i^\top\left(\dpart{y} E\right)^{-1}\bigr)_k \; (y_i'(\overbar{p}))_\ell \cdot \ddpart{y_i}{y_j} E_k \cdot (y_j'(\overbar{p}))_m}  \right. \\
& + 2 \sum\limits_{\ell, m = 1}^{n_p} \sum\limits_{k = 1}^{n_y} \enbrace{\sum\limits_{j = 1}^{n_y} \bigl(\dpart{y}h_i^\top\left(\dpart{y} E\right)^{-1}\bigr)_k \; \ddpart{p_\ell}{y_j} E_k \cdot (y_j'(\overbar{p}))_m} \\
& + \left. \sum\limits_{\ell, m = 1}^{n_p} \sum\limits_{k = 1}^{n_y} \bigl(\dpart{y}h_i^\top\left(\dpart{y} E\right)^{-1}\bigr)_k \; \ddpart{p_\ell}{p_m} E_k \right] e_\ell \otimes e_m,
\end{align*}
whereby $ \otimes $ denotes the standard tensor product and with dropped dependencies on \linebreak $ h = h(y(\overbar{p}), \overbar{p}, q) $ and $ E = E(y(\overbar{p}), \overbar{p}, q) $ for the sake of clarity. Altogether, $H(\Omega)$, $J(\Omega)$ and therefore also $C(\overbar{p}, \Omega)$ can be determined using the expressions given in 
\eqref{eq:partial_r}--\eqref{secondDerStateEq}.

\end{document}